\pdfoutput=1

\documentclass[11pt]{article}

\usepackage[]{acl}

\usepackage{times}
\usepackage{latexsym}

\usepackage[T1]{fontenc}

\usepackage[utf8]{inputenc}

\usepackage{microtype}
\usepackage{float}
\usepackage{amsmath}
\usepackage{amsfonts}
\usepackage{multirow}
\usepackage{makecell}
\usepackage{array}
\usepackage{hhline}
\usepackage{booktabs}
\usepackage{graphicx}
\usepackage{subfig}
\usepackage{xcolor}
\usepackage{soul}
\usepackage{adjustbox}

\newcommand{\RNum}[1]{\uppercase\expandafter{\romannumeral #1\relax}}
\newcommand{\modelname}[0]{\textsc{EntFA}}

\newcommand{\datasetname}[0]{\textsc{XEnt}}

\definecolor{darkblue}{HTML}{033394}
\definecolor{darkgreen}{HTML}{005e19}
\definecolor{darkred}{HTML}{8a0000}
\definecolor{darkyellow}{HTML}{a37800}

\title{Hallucinated but Factual! Inspecting the Factuality of Hallucinations in Abstractive Summarization}

\author{Meng Cao \qquad Yue Dong \qquad Jackie Chi Kit Cheung \\ \\
    School of Computer Science, McGill University, Montreal, QC, Canada \\
    MILA, Montreal, QC, Canada \\ \\
    {\small \{\tt meng.cao@mail, yue.dong2@mail, jcheung@cs\}.mcgill.ca}}

\begin{document}
\maketitle
\begin{abstract}
State-of-the-art abstractive summarization systems often generate \emph{hallucinations}; i.e., content that is not directly inferable from the source text. Despite being assumed incorrect, we find that much hallucinated content is factual, namely consistent with world knowledge. These factual hallucinations  can be beneficial in a summary by providing useful background information. 
In this work, we propose a novel detection approach that separates factual from non-factual hallucinations of entities. 
Our method utilizes an entity's prior and posterior probabilities according to pre-trained and finetuned masked language models, respectively. Empirical results suggest that our approach vastly outperforms two baselines 
and  strongly correlates with human judgments. 
Furthermore, we show that our detector, when used as a reward signal in an off-line reinforcement learning (RL) algorithm, significantly improves the factuality of summaries while maintaining the level of abstractiveness.
\footnote{\url{https://github.com/mcao516/EntFA}}

\end{abstract}

\section{Introduction}
State-of-the-art abstractive summarization systems can generate fluent summaries with high automatic evaluation scores in terms of ROUGE \citep{lin-2004-rouge}. However, recent studies have shown that these systems are prone to hallucinate content that is not supported by the source document \cite{maynez-etal-2020-faithfulness, kang-hashimoto-2020-improved, durmus-etal-2020-feqa, zhao-etal-2020-reducing, filippova-2020-controlled, kryscinski-etal-2020-evaluating}. For instance, \citet{maynez-etal-2020-faithfulness} discovered that 64.1\% of the summaries generated by a BERT-based abstractive summarization model on \textsc{XSum} \citep{narayan-etal-2018-dont} contain hallucinations.

Previous studies commonly assume that hallucination is an undesirable behavior in abstractive summarization systems. They investigate the cause of model hallucination \cite{kang-hashimoto-2020-improved, wang-sennrich-2020-exposure} and propose methods that reduce the frequency of all hallucinations \cite{filippova-2020-controlled, zhao-etal-2020-reducing, nan-etal-2021-entity, narayan2021planning}.

Our stance in this paper is that \textit{ hallucinations are not always undesirable}: 
many factual hallucinations provide additional world knowledge that is important for summary comprehension. Table~\ref{table:hallucination_example} presents one such example from \textsc{XSum}: the hallucinated content \textit{European Commission President} provides additional background information on the role of \textit{Mr. Juncker}. Figure~\ref{figure:Venn} illustrates our proposed view of the relationship between the contents of a summary, of source documents and world knowledge. Factual hallucinations refer to content that is verifiable by world knowledge but not inferable from source text.

\begin{table}[t]
\small
\renewcommand{\arraystretch}{1}
\setlength\tabcolsep{2.5pt}
\setlength{\belowcaptionskip}{-0.6cm}
\centering
\begin{tabular}{|p{7.3cm}|}
  \hline
  {\bf Source}: \\
    Under the proposals, 120,000 additional asylum seekers will be distributed among EU nations, with binding quotas. (...) {\bf Mr Juncker} told the European Parliament it was ``not a time to take fright''. (...) He said tackling the crisis was ``a matter of humanity and human dignity''. ``It is true that Europe cannot house all the misery in the world. But we have to put it into perspective.'' (...) \\
  \hline
  {\bf Generation}: \\
  \hl{European Commission President} \hl{Jean-Claude} Juncker has set out his proposals for dealing with the migrant crisis in a speech to MEPs, saying Europe cannot house all the misery in the world. \\
  \hline
\end{tabular}
\caption{\label{table:hallucination_example} Example of factual hallucinations in a BART generated summary on \textsc{XSum}. Both the title ``European Commission President'' and the first name ``Jean-Claude'' is not mentioned in the document but factual.}
\end{table}

\begin{figure}[t!]
\centering
\setlength{\belowcaptionskip}{-0.2cm}
\includegraphics[scale=0.6]{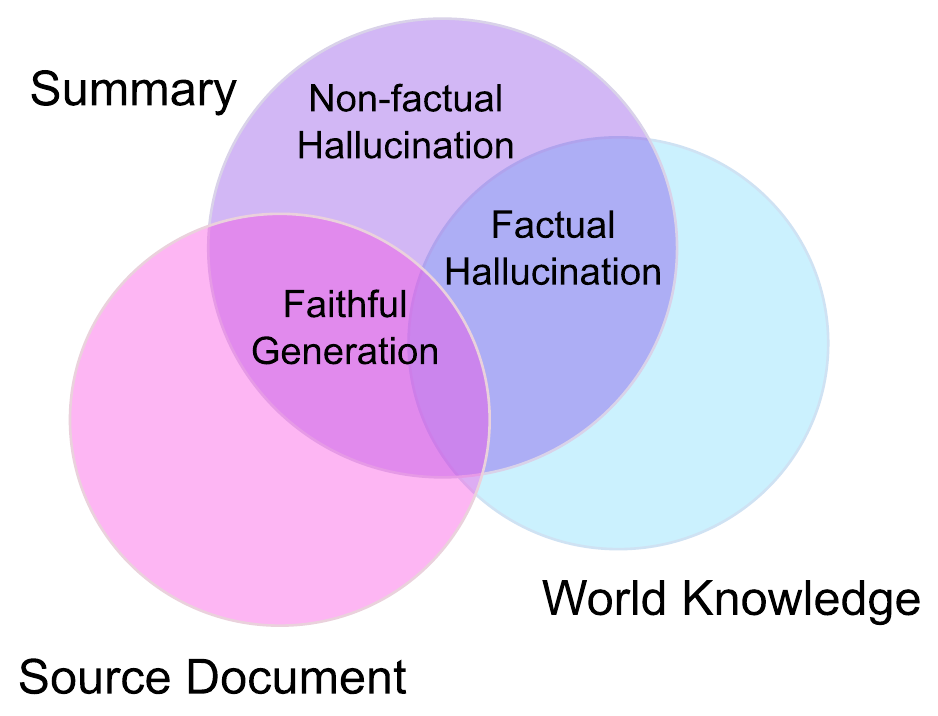}
\caption{The relationship between faithful generation, factual/non-factual hallucination, source document and world knowledge.}
\label{figure:Venn}
\end{figure}


We thus argue that not all hallucinations should be treated equally; in particular, factual hallucinations may be less deleterious or even potentially beneficial to to be included in a summary, as opposed to non-factual ones. We propose a method to classify entities according to whether they are hallucinations and whether they are factual (if hallucinated). We focus on entities (e.g., persons, locations, dates, cardinal numbers) because they are necessary to express the most salient pieces of information in a summary. Moreover, entity hallucinations are  common in generated summaries. As we will show later in our work, about 30\% of entities generated by BART \cite{lewis-etal-2020-bart} on \textsc{XSum} test set are hallucinated.

Our approach is inspired by the observation that many hallucinated entities are generated with low probabilities. 
This observation suggests that the summarization model's confidence correlates with the factuality statuses of generated entities. In other words,  the uncertainty 
is indicative of the likelihood of whether generated entities are hallucinated and non-factual. 

We refer to the probability of an entity being in a summary without considering the source document as its prior probability, and its probability given the document as its posterior probability.
Our assumption is that if an entity in a generated summary results in a factual error, giving the source should not provide more evidence for it, resulting in a small change in probability between the prior and the posterior. Based on this assumption, we propose to use the prior and posterior probabilities as the key features in a simple classifier that predicts an entity's hallucination status and factuality.

Due to the lack of fine-grained hallucination annotation, we create an entity-level hallucination and factuality annotation on the \textsc{XSum} dataset.
We evaluate our detection method on this annotated dataset as well as annotations from \citet{maynez-etal-2020-faithfulness}. On both datasets, our approach outperforms two baseline models at identifying non-factual hallucinations. 
In addition, our approach has a strong correlation with the factuality scores given by human judges. Besides, we show that our detector, when used as a reward signal in training neural-based summarizers with the off-line RL algorithm, significantly improves the factuality of generated summaries even when the underlying dataset is noisy.

Our contributions are the following:
(\textit{i}) We demonstrate that an entity's prior and posterior probabilities can be used to infer whether it is hallucinated and factual. Based on this hypothesis, we propose a novel approach for entity-level hallucination detection and factuality checking. Our approach outperforms two baselines from previous work on two human-annotated datasets, in addition to having a strong correlation with summary-level factuality scores given by human judges. 
(\textit{ii}) We empirically demonstrate that our classifier can provide reliable reward signals for RL algorithms, leading to improved factuality while maintaining the level of abstractiveness in generated summaries.
(\textit{iii}) We create a set of entity-level hallucination annotations. 

\section{Related Work}
The correctness of summarization systems' outputs has been evaluated as one aspect of content selection in the past, for example using the Pyramid method \cite{nenkova2004evaluating}. As neural abstractive summarizers have become popular, their issues with correctness have sparked much recent work that focus specifically on model hallucinations and summary factuality \cite{kryscinski-etal-2020-evaluating}.

\subsection{Model Hallucination}
\citet{maynez-etal-2020-faithfulness} conducted
a large-scale human evaluation of several neural abstractive summarization systems, and found that hallucinations are common among the outputs of different summarization models.


Recently, many methods have been proposed to reduce model hallucination. \citet{kang-hashimoto-2020-improved} 
propose a ``loss truncation'' training algorithm that filters out noisy training samples which may lead a model to hallucinate.
\citet{zhao-etal-2020-reducing} use a verification system to recognize non-factual quantities in summaries and adopt a re-ranking system to reduce the number of hallucinated quantities in the final output summary. \citet{narayan2021planning} use entity chains to mitigate the hallucination problem in the generation of abstractive summaries. \citet{nan-etal-2021-entity} show that data filtering and use a summary-worthy entity classification task as an auxiliary training objective can help improve model's entity-level factuality.

\citet{filippova-2020-controlled} proposed a method for controlling hallucination in data-to-text generation task. They suggest that a conditional language model (CLM) will put more probability mass on a non-hallucinated entity than an unconditional language model (LM). Our work differs in that we focus on both hallucination and factuality. Also, our method works at the entity-level rather than the sentence-level, and is geared towards text summarization.


\subsection{Summary Factuality}
Another line of work focuses on evaluating the factual consistency of abstractive summarization systems. 
\citet{kryscinski-etal-2020-evaluating} train models on an artificially corrupted dataset for factual errors detection. \citet{cao-etal-2020-factual} induce artificial perturbations in text to train a summary error correction system, but find that there is a large gap between such artificial perturbations and the type of hallucinations that are generated by abstractive summarizers. \cite{goyal-durrett-2020-evaluating} measure factual consistency by checking whether the semantic relationship manifested by individual dependency arcs in the generated summary is supported by the source document.
\citet{wang-etal-2020-asking, dong-etal-2020-multi-fact, durmus-etal-2020-feqa} measure and improve the factual consistency of summaries by asking and answering questions based on generated summaries and input documents. 

\section{Method}
In this section, we propose a novel detection approach that separates factual from non-factual hallucinations of entities (Section~\ref{sec:method-classifier}), and present
a factuality-aware training framework for summarization models trained on noisy dataset (Section~\ref{sec:rl_factuality_reward}). 

\subsection{Problem Statement}
Let $(S, R)$ be a pair of a source document and a reference summary, where $S = (s_1, ... , s_M)$ is the source document with $M$ tokens, and $R = (r_1, ... , r_L)$ is the reference summary with $L$ tokens. Let $G = (g_1, ... , g_N)$ be the model-generated summary with $N$ tokens.
For each named entity $e_k$, which we assume to be a span of tokens $g_{i_k}, ... , g_{i_k+|e_k|-1}$ $(|e_k| \geq 1)$ starting at position $i_k$ in $G$, the task is to determine whether $e_k$ is hallucinated, and whether it is factual. We define an entity as hallucinated if it is not directly inferable in its generated context given the input document $S$. If an entity is hallucinated, we further classify it into two subtypes: \textit{factual hallucinations} and \textit{non-factual hallucinations}. Factual hallucinations cannot be directly entailed from the source document but are factually correct based on world knowledge (see Table \ref{table:hallucination_example}). Non-factual hallucinations are entities that are neither inferable from the source nor factual.

\subsection{The Prior \& Posterior Probability of an Entity}
\label{sec:method-classifier}
We now define the prior and posterior probabilities of an entity, which we will use to predict its hallucination and factuality statuses.

For entity $e_k$, we define its prior probability $p_{\textrm{prior}}(e_k)$ as the probability of its generation by a language model that does not have access to the source text. If $e_k$ spans multiple tokens, we compute its probability auto-regressively. Let $c_k$ be the context of entity $e_k$ in $G$, excluding the tokens in $e_k$. Then:
\begin{align}
p_{\textrm{prior}}(e_k) &= P_{\textrm{MLM}}(e_k \ | \ c_k)\\
 &= \prod_{t=1}^{|e_k|} P_{\textrm{MLM}}(e_k^t \ | \ e_k^{1...t-1}, c_k)
\end{align}
which we compute using a masked language model $P_{\textrm{MLM}}$.


The posterior probability $p_{\textrm{pos}}(e_k)$ of entity $e_k$ is the conditional probability of the entity given the context and the source text: 

\begin{align}
p_{\textrm{pos}}(e_k) &=  P_{\textrm{CMLM}}(e_k \ | \ c_k, S) \\  
&= \prod_{t=1}^{|e_k|} P_{\textrm{CMLM}}(e_k^t \ | \ e_k^{1...t-1}, c_k, S),
\end{align}
where CMLM is a conditional masked language model.
CMLM is an encoder-decoder model that is trained with a masked language model objective on a parallel dataset. Specifically, a CMLM predicts a target sequence $T$ given a source text $S$ and part of the target $T_{\textrm{masked}}$, where $T_{\textrm{masked}}$ is the target sequence with a random entity being masked. In order to correctly generate the missing part of the sentence, the model needs to condition on both $T_{\textrm{masked}}$ and $S$. 
Alternatively, we can calculate the entity's posterior probability using a conditional language model (CLM) instead of a CMLM.
In this case, the entity's posterior probability is defined as $P_{\textrm{CLM}}(e_k \ | \ c_{e_k}, S)$ where $c_{e_k} = g_1, ..., g_{i-1}$. Note that CLM is only conditioned on the left context.

\paragraph{Training a Discriminator}
\label{sec:knn}
To classify the hallucination and factuality statuses of a given entity, we need to train a discriminator model. We use the K-Nearest Neighbors (KNN) algorithm since it requires no training and makes minimal assumptions about the form of the decision boundary, as a non-parametric method.
It also offers adequate interpretability. The KNN classifier is trained using the prior and posterior probabilities as features on our labeled dataset. Since the classifier is used for entity hallucination and factuality assessment, we refer to it as {\bf \modelname}.
Besides using the prior/posterior probability as features, we also add a binary overlap feature that indicates whether the entity appears in the document. 
We train two classifiers for hallucination detection and factuality checking tasks respectively. 

\subsection{Improving the Factuality of Abstractive Summarization Systems}
\label{sec:rl_factuality_reward}
We now propose a factuality-aware training approach for summarization systems that combines our factuality assessment model with the latest off-line RL technique.

\paragraph{RL for Text Generation}
Sequence generation of the tokens in the summary text can be viewed as a finite Markov Decision Process (MDP). At each time-step $t$, the state $s_t$ consists of the source text $x$ and the previously generated tokens $y_{<t}$, $s_t=(y_{<t}, x)$. The agent, which is the summarization model, takes an action by generating a new token $a_t$. Depending on the action taken, the agent gets a reward $r_t = R(s_t, a_t)$ and deterministically transitions to the next state $s_{t+1}=(y_{<t+1}, x)$. The probability of each action (i.e., token) is specified by the policy $\pi_{\theta}(a_t | s_t)$. The goal of the agent is to maximize the discounted cumulative reward throughout the trajectory: $J(\theta) = \mathbb{E}_{\tau \sim \pi} \Big[ \sum_{t=0}^T \gamma^t r_t \Big]$.

When training the summarization model with human-written reference summaries, we can frame the training process as an off-line RL problem with expert demonstrations (i.e., the reference summaries). In this setting, since we are sampling trajectories from a behavior policy, we need an importance sampling term $w_t$ to correct the gradient estimation. Following \citet{pang2021text}'s work, we approximate $w_t$ with $\pi_{\theta}(a_t|s_t)$ and this gives us the following objective:
\begin{equation}
\begin{split}
& \nabla_{\theta} J(\theta) = \\
& \mathbb{E}_{\tau \sim \pi_b} \Big[ \sum_{t=0} \pi_{\theta}(a_t|s_t) \nabla_{\theta} \log \pi_{\theta} (a_t \; | \; s_t) \hat{Q}(a_t, s_t) \Big]
\end{split}
\end{equation}
where $\hat{Q}(a_t, s_t) = \sum_{t'=t}^T \gamma^{t'-t}r_{t'}$ is the estimated return from state $s_t$ and $\pi_b$ is the behavior policy from which we draw trajectories $\tau$. In our case, $\pi_b$ is the (noisy) summarization dataset.

\paragraph{Training with a Factuality-based Reward} 
\label{sec:factuality_reward}
One problem in the off-line RL setting is that expert demonstrations, which in our case are the reference summaries, are often noisy and contain content that cannot be inferred from the source document. The commonly used teacher forcing training encourages the model to blindly imitate the training data, which leads to model hallucination at inference time \cite{kang-hashimoto-2020-improved}.

To discourage the model from overfitting to the noise in the training set, we use the predictions from our classifier as factuality reward signals to guide the training of the summarization model. In the off-policy learning stage, we use our factuality classifier to label all the entities in the training set. 
If an entity is classified by our classifier as ``non-factual'', we consider it noise and give it a negative reward $-r_{\textrm{nfe}}$. For factual entities and other tokens, we use the posterior probability from a MLE-trained model as token-level rewards, as in \cite{pang2021text}. Formally, we have:
\begin{equation*}
R(s_t, a_t) = \
\begin{cases}
    -r_{\textrm{nfe}}, & \text{if $a_t$ is non-factual} \\
    p_{\textrm{MLE}}(a_t|s_t), & \text{otherwise} 
\end{cases}
\end{equation*}

\section{Evaluation Tasks and Datasets}
In this section, we first discuss the datasets used for the evaluation of {\modelname}. Then, we introduce the evaluation tools used for evaluating the effectiveness of our factuality-aware training method.

\subsection{Hallucination and Factuality Assessment}
\paragraph{\datasetname{} dataset} To study entity hallucination and factuality in abstractive summarization, we need annotations of entity- or token-level hallucination. To the best of our knowledge, there is no such dataset available. Therefore, we create a dataset ourselves, which we call the {\datasetname} dataset.

We\footnote{Two coauthors and three graduate students.} annotate 800 summaries generated by BART, which is one of the current state-of-the-art abstractive summarization models.
The input documents are randomly selected from \textsc{XSum} test set.
We choose \textsc{XSum} because it is more abstractive than other summarization datasets.
We extract 2,838 entities from the 800 generated summaries. We randomly select 30\% of the samples as our test set.

We manually labeled each entity with one of the following three tags: non-hallucinated, factual hallucination, and non-factual hallucination. First, we check whether the entity can be directly entailed using the information from the source document. If so, then the entity is non-hallucinated; otherwise, we need to decide whether the entity is factual using world knowledge. This often requires external resources such as Wikipedia or Google Search. Based on the search result, the entity is labeled as either factual hallucination or non-factual hallucination. If there is no information found online to prove or disprove the hallucinated entity, it is labeled as non-factual. There is a special case where the entity misrepresents information from the document. For instance, the summary might include a number from the document but that number is actually related to a different event. In this case, the entity is considered as an intrinsic hallucination \cite{maynez-etal-2020-faithfulness}. In this work, we will focus on extrinsic hallucinations, so we discarded all intrinsic hallucinations in our experiments. Table~\ref{table:dataset} shows the distribution of entities by hallucination and factuality status in our labeled dataset. We show an example for each hallucination type in Appendix~\ref{sec:appeddix_example}.

\paragraph{Inter-Annotator Agreement}
We report Fleiss's Kappa ($\kappa$) to access the reliability of agreement between annotators. We compute agreement on 800 annotated entities and obtain almost perfect agreement ($0.80 \leq \kappa \leq 1.00$) with $\kappa = 0.809$. Following \citet{pagnoni-2021-frank}, we also report the percentage $\mu$ of annotators that agree with the majority class. We  obtain $\mu = 0.931$ of annotators agreeing with the majority class on the four-category annotation which shows substantial agreement.

\paragraph{\textsc{MEnt} Dataset}
\label{para:convert}
Recently, \citet{maynez-etal-2020-faithfulness} released a set of factuality and hallucination annotations for \textsc{XSum}. For each generated summary, they labeled the hallucinated spans as well as the overall factuality of the summary. 
Compared with our labeling approach, their annotation has a lower granularity and does not distinguish between factual hallucination and non-factual hallucination. Therefore, we have to convert their dataset first before using it for evaluation.

To perform entity-level factuality checking on their dataset, we do the following: First, we extract entities from the annotated summaries. For entities that are extracted from factual summaries, we label them as factual entities. For each entity from non-factual summary, if it is inside an extrinsic hallucinated span, then we assume the entity is non-factual. Otherwise the entity is labeled as a factual. This process gives us a new dataset that has the same format as ours for entity-level factuality evaluation. We refer to this new dataset as the \textsc{MEnt} dataset. 

However, it is worth pointing out that the converted dataset is noisy. For instance, in \citet{maynez-etal-2020-faithfulness}'s annotation, the entire generated summary is often labeled as a hallucinated span if it does not capture the meaning of the document well. In this case, the hallucinated span could still contain faithful entities with respect to the source document. This could result in false-positive non-factual entities after the conversion. Therefore, we filter out entities in the extrinsic hallucination span that also appear in the source document.

\subsection{Correlation with Human Judgments of Factuality}
In addition to entity-level classification performance, we also evaluate our methods by correlating them against human judgments of factuality. Previous work has collected summary-level judgments of factuality from human annotators, which are then correlated with automatic evaluation measures applied to those summaries. To apply our entity-level method, we use the lowest classifier confidence for the factual class among its entities as the factuality score for the entire summary. We evaluate correlation on two datasets by \citet{pagnoni-2021-frank} and \citet{wang-etal-2020-asking}. 
\subsection{Evaluating the Factuality of Summarization Systems}
To evaluate our factuality-aware training approach proposed in Section~\ref{sec:rl_factuality_reward}, we train a summarization model with factuality rewards and evaluate model's predictions on \textsc{XSum} test set. 
To evaluate the faithfulness of generated summaries,
we use automatic faithfulness evaluation tools FEQA \cite{durmus-etal-2020-feqa} and DAE \cite{goyal-durrett-2020-evaluating}\footnote{In this work, we define the faithfulness of the summary as whether it is faithful with respect to the source. Factuality as whether is factual with respect to world knowledge.}. We also calculate ROUGE scores, and the percentage of $n$-grams and percentage of entities in the generated summaries that are not found in the source document (ENFS). The percentage of novel $n$-grams reflects the extractiveness of summarization model.

\section{Experiments}
\paragraph{Training CMLM \& MLM}
\label{sec:train_cmlm}
For training the CMLM, we use both \textsc{XSum}, \citet{xsum-emnlp}) and the CNN/Dailymail dataset \citep{NIPS2015_afdec700} dataset. 
To build a training corpus for CMLM, we randomly select one entity in each reference summary and mask it with a special $[\texttt{MASK}]$ token. 
We append a $[\texttt{S}]$ token at the beginning of each summary. The document and summary are concatenated together (separated by $[\verb|\|\texttt{S}]$ token) as CMLM's input. The training target is the reference summary without any masking. If there is no specification, we use the CMLM trained on \textsc{XSum}.
For the MLM, we use the large BART model. BART is pre-trained on five different reconstruction tasks including token masking and text infilling.
For more experimental setup and hyper-parameter setting details, see Appendix \ref{sec:hyperparameters}.

\begin{table}[t!]
\vspace{-1.8mm}
\renewcommand{\arraystretch}{0.9}
\setlength{\belowcaptionskip}{-0.3cm}
\begin{center}
\begin{adjustbox}{width=\columnwidth, center}
\begin{tabular}{l|c|c}
\toprule
\bf Label & \bf \#Samples & \bf Total Ent. \\ 
\midrule
Non-hallucinated & 1,921 (67.69\%) & \multirow{4}{*}{2,838} \\ 
Factual hal. & 441 (15.54\%) & ~ \\ 
Non-factual hal. & 421 (14.83\%) & ~ \\ 
Intrinsic hal. & 55 (1.94\%) & ~ \\ 
\bottomrule
\end{tabular}
\end{adjustbox}
\end{center}
\caption{\label{table:dataset} Statistics of labeled dataset. Entities are extracted from BART generated summaries on \textsc{XSum}.} 
\end{table}


\begin{table}[t!]
\vspace{-1.8mm}
\renewcommand{\arraystretch}{0.9}
\setlength{\belowcaptionskip}{-0.5cm}
\begin{adjustbox}{width=\columnwidth, center}
\begin{tabular}{l|cc|cc}
\toprule
\multirow{2}{*}{~} & \multicolumn{2}{c|}{\bf Hallucination} & \multicolumn{2}{c}{\bf Factuality} \\ 
~ & Acc. & F1 & Acc. & F1 \\ 
\midrule
Word overlap & 92.93 & 91.73 & 81.25 & 74.19 \\
LM-based & 74.18 & 54.99 & 84.54 & 57.80 \\
\midrule
{\modelname} (ours) & \bf 93.09 & \bf 91.91 & \bf 90.95 & \bf 81.82 \\
\bottomrule
\end{tabular}
\end{adjustbox}
\caption{\label{tab:evaluation} Entity's factuality and hallucination status evaluation results on {\datasetname}. We report the accuracy and (macro) F1 score on the test set. The number of neighbors $k$ is set to 20 for both tasks.}
\end{table}

\subsection{Classification Experiments}
\label{sec:main}
\paragraph{Baselines} Our baseline models are based on the two methods proposed by \citet{filippova-2020-controlled}: the \emph{overlap-based} method and the \emph{LM-based} method. The \emph{overlap-based} method checks the word overlap between the summary and the source document. In our case, we check whether a given entity in the generated summary also exist in the source document. If it does not, the entity is classified as both hallucinated and non-factual. The \emph{LM-based} method uses LM and CLM to compute the token's prior and posterior probability.
In \citet{filippova-2020-controlled}'s work, they compare the value of $p_{\textrm{prior}}$ and $p_{\textrm{pos}}$. If the generated token does not match the reference and $p_{\textrm{prior}}$ is greater than $p_{\textrm{pos}}$, the token is classified as hallucinated. Since we are evaluating the generated summary but not the reference, we modify their method to the following: if the entity is not found in the source and $p_{\textrm{prior}} > p_{\textrm{pos}}$, then the entity is classified as non-factual and hallucinated. 

\paragraph{Evaluation Results on {\datasetname}}
Table~\ref{tab:evaluation} shows the evaluation results of our classifiers and baselines in terms of both entity factuality and hallucination status classification.
The results show that our approach outperforms two baselines by large margins on the factuality classification task. 
To show that our model is statistically better than the baselines, we run a $10$-fold cross-validated paired t-test comparing our model with two baselines. 
The results show that our model is better than the baseline models with $p$-value less than $3.3e-5$.
On the hallucination detection task, the word-overlap baseline achieves a relatively high accuracy 92.93\% compared with our model's 93.09\%. However, the word-overlap model alone cannot distinguish between factual and non-factual hallucinations. This is the reason for its performance degradation on factuality classification task.

For hallucination classification, the reason computing word overlap with the source does not completely solve the hallucination detection problem is that hallucination is defined based on the semantic relationship between the source and the summary. There can exist words that are not in the source document but which can nevertheless be inferred from it. 

\paragraph{Evaluation Results on \textsc{MEnt} Dataset}
\label{sec:mentx}

\begin{table}[t!]
\renewcommand{\arraystretch}{0.9}
\setlength{\belowcaptionskip}{-0.1cm}
\begin{center}
\begin{tabular}{l|cc}
\toprule
~ & \bf Acc. & \bf F1 \\ 
\midrule
Word overlap & 68.22 & 54.68 \\
LM-based & 67.48 & 48.02 \\
\midrule
{\modelname} (ours) & \bf 78.48 & \bf 60.23 \\
\bottomrule
\end{tabular}
\end{center}
\caption{\label{tab:evaluation_google} Entity-level factuality evaluation results on converted \textsc{MEnt} Dataset (\citet{maynez-etal-2020-faithfulness}).}
\end{table}


Table~\ref{tab:evaluation_google} shows the evaluation results on \textsc{MEnt}. {\modelname} are learned on our annotated training set with $k$ set to 20.
The performance of all models is lower on this dataset. This may be due to fact that the converted dataset is noisier than the {\datasetname} dataset (see Section \ref{para:convert}).
For the factuality classification task, our model outperforms two baseline models. This demonstrates the generalizability of our approach.

\subsection{Correlation Experiments}
\label{sec:corr}
\begin{table}[t!]
\small
\renewcommand{\arraystretch}{1.0}
\setlength{\belowcaptionskip}{-0.1cm}
\begin{center}
\begin{adjustbox}{width=\columnwidth, center}
\begin{tabular}{c|c|c}
\toprule
Metric & \makecell{FRANK \\ ({\small Partial Pearson's $\rho$})} & \makecell{Wang et al. \\ (PCC)} \\
\midrule
BLUE & 0.139 & 0.118 \\
ROUGE-1 & 0.155 & 0.132 \\
ROUGE-L & 0.156 & 0.089 \\
METEOR & 0.155 & 0.100 \\
BERTScore & -0.0359 & 0.025 \\
QAGS & -0.0225 & 0.175 \\
FEQA & 0.0242 & - \\
DAE & 0.0444 & - \\
\midrule
{\modelname} (ours) & \bf 0.183 & \bf 0.268 \\
\bottomrule
\end{tabular}
\end{adjustbox}
\end{center}
\caption{\label{tab:correlation} Summary-level Pearson correlation coefficients between various automatic metrics and human judgments of factuality for \textsc{\textsc{XSum}} datasets. In the middle column, we use the FRANK benchmark for factuality evaluation metrics from \citet{pagnoni-2021-frank}; In the right column, we use the human judgments collected by \citet{wang-etal-2020-asking}. All baselines' coefficient values are cited from their papers. }
\end{table}
Table~\ref{tab:correlation} presents the correlation evaluation results. On \citet{pagnoni-2021-frank}'s benchmark dataset, our approach has the highest partial Pearson correlation coefficient $\rho =$ 0.183 ($p < 1e^{-8}$). On \citet{wang-etal-2020-asking}'s dataset (right column), our approach outperforms all other automatic metrics significantly. These results indicate that our model can be used for automatic factuality evaluation of summaries at both the entity and sentence levels.

\subsection{Factuality Evaluation Results of Summarization Systems}
\begin{table*}[t!]
\vspace{-4mm}
\renewcommand{\arraystretch}{0.90}
\centering
\setlength{\belowcaptionskip}{-0.1cm}
\begin{adjustbox}{width=\columnwidth*2, center}
\begin{tabular}{l|cc|cc|ccc|cc}
\toprule
\multirow{2}{*}{~}  & \multicolumn{2}{c|}{ROUGE} & \multicolumn{2}{c|}{\% of novel n-gram} & \multicolumn{3}{c|}{Faithfulness} & \multicolumn{2}{c}{\modelname} \\ 
System & R1 $\uparrow$ & RL $\uparrow$ & {\small unigrams} $\uparrow$ & {\small bigrams} $\uparrow$ & \% ENFS $\downarrow$ & FEQA $\uparrow$ & DAE $\uparrow$ & {\small \% Factual Ent $\uparrow$} & {\small \% Factual Hal $\uparrow$} \\ 
\midrule
MLE & 45.1 & 37.3 & 27.86 & 74.47 & 42.0 & 25.9 & 34.6 & 82.8 & 21.4 \\
RL  & \bf 45.8 & \bf 37.6 & 28.14 & 74.73 & 43.2 & 25.6  & 33.3 & 82.8 & 21.6 \\
LM-based & 43.2 & 34.6 & \bf 29.75 & \bf 75.86 & 38.2 & 24.2  & 31.3 & 87.4 & 21.7 \\
\midrule
Loss trunc (c=0.3) & 44.1 & 36.0 & 26.82 & 73.39 & 41.3 & 26.3  & 36.4 & 83.9 & 21.3 \\
Loss trunc (c=0.7) & 42.7 & 34.8 & 26.61 & 73.19 & 40.6 & 26.7  & 38.8 & 84.1 & 20.7 \\
\midrule
{Ours ($r_{\textrm{nfe}}=2.0$)} & 44.6 & 36.2 & 27.71 & 74.90 & 37.2 & 26.5 & 37.3 & 90.1 & \bf 24.0 \\
{Ours ($r_{\textrm{nfe}}=4.0$)} & 43.0 & 34.9 & 26.87 & 74.11 & \bf 32.8 & \bf 27.3 & \bf 40.8 & \bf 92.5 & 22.4 \\
\bottomrule
\end{tabular}
\end{adjustbox}
\caption{\label{tab:summarization_eval} Comparison of different summarization models. Results are evaluated on \textsc{XSum}'s official test set. ``\% Factual Ent'' and ``\% Factual Hal'' are the percentage of factual entities and factual hallucinations classified by {\modelname} model respectively. ``\% ENFS'' is the percentage of entities in generated summary that not found in source document. For the loss truncation baseline, $c$ is the percentage of data to be dropped.}
\vspace{-3mm}
\end{table*}

\paragraph{Baselines} We compare our approach with four baselines: a teacher forcing-based summarizer (MLE), a RL-based summarizer (RL) \cite{pang2021text} and a summarizer trained with the loss truncation technique from \citet{kang-hashimoto-2020-improved}. We also replace our factuality assessment model {\modelname} with \citet{filippova-2020-controlled}'s approach (LM-based) for entity factuality labeling as another baseline model (see Section~\ref{sec:factuality_reward}).

Table~\ref{tab:summarization_eval} shows the evaluation results on \textsc{XSum}. The results show that our approach outperforms all baselines with fewer non-factual entities and higher faithfulness scores. Note that our approach has the lowest ENFS rate while having the highest percentage of factual hallucinations. Compared with the loss truncation baseline, our method also produces more novel $n$-grams. These show that our method does not improve the factuality of the model by simply making the model more extractive.

\begin{figure}[t!]
\centering
\vspace{-1mm}
\setlength{\belowcaptionskip}{-0.1cm}
\includegraphics[scale=0.4]{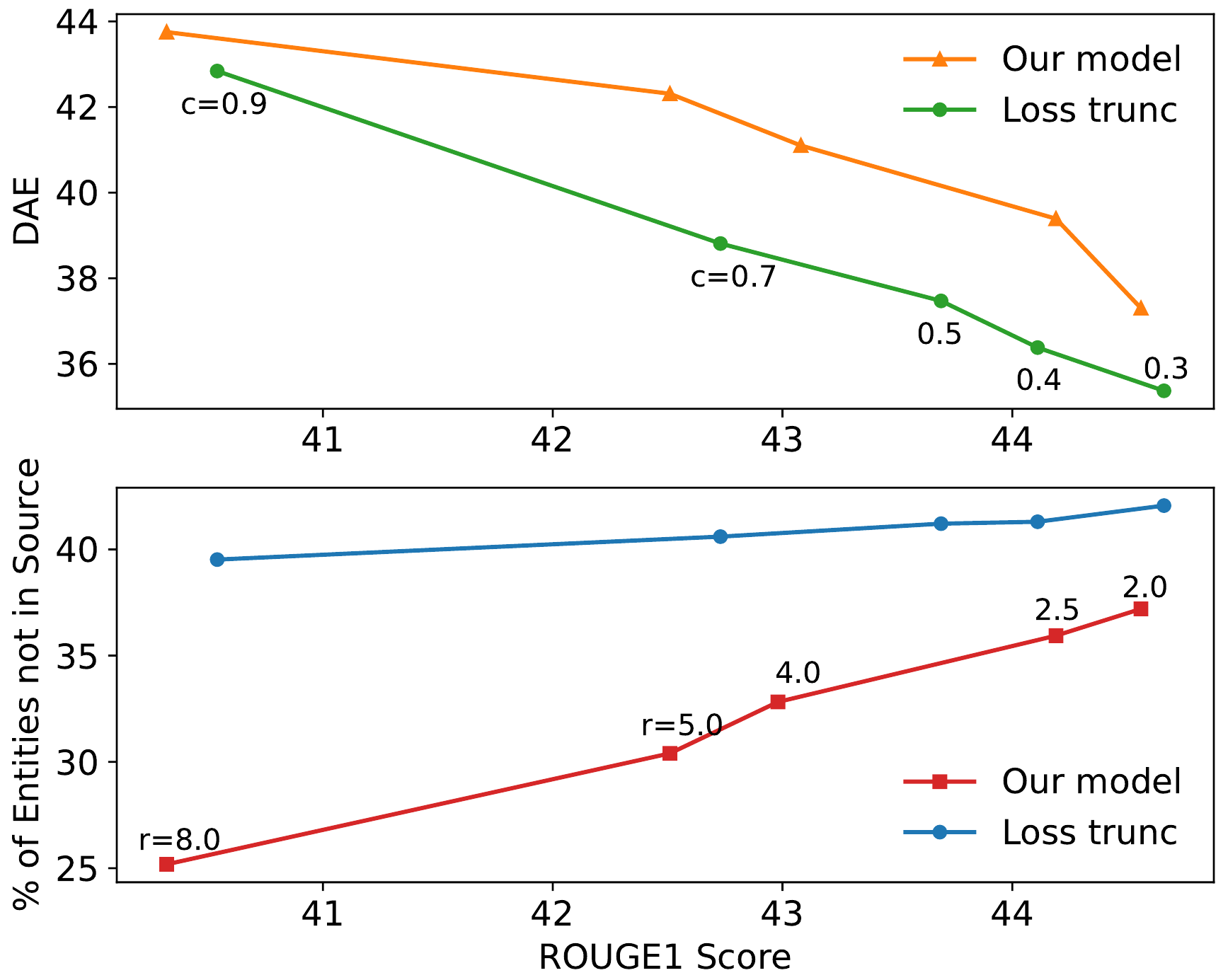}
\caption{The factuality and ROUGE score trade-off curve on \textsc{XSum}. We use different reward value $r_{\text{nfe}}$ for our approach and different drop rate $c$ for the loss truncation baseline. }
\label{figure:dae_nefs}
\end{figure}

Figure~\ref{figure:dae_nefs} shows the factuality and abstractiveness trade-off curves of our model compared to the loss truncation baseline. At the same level of ROUGE performance, our method can obtain a higher factuality score. This further proves that our model can generate both factual and high-quality summaries compared with the loss truncation baseline.

\section{Analysis}

\subsection{Ablation Studies}
\begin{table}[t!]
\small
\renewcommand{\arraystretch}{0.9}
\setlength{\belowcaptionskip}{-0.47cm}
\begin{center}
\begin{tabular}{l|cc}
\toprule
~ & \bf Factuality & \bf Hallucination \\ 
\midrule
{\modelname} & 81.82 & 91.91 \\
\midrule
w/o overlap & 77.18 & 74.83 \\
w/o prior & 80.12 & 91.32 \\
w/o posterior & 70.30 & 91.12 \\

\bottomrule
\end{tabular}
\end{center}
\caption{\label{tab:ablation} Ablation studies of different feature combination. We report the F1 score on {\datasetname} test set.}
\end{table}

To explore the effect of each feature, we conduct an ablation study by training the KNN classifier with fewer features. The results are illustrated in Table~\ref{tab:ablation} and show that all the proposed features are useful.
For factuality classification, The performance w/o posterior drops significantly from 90.95 to 85.69.
This result suggests that the posterior probability is crucial for factuality classification. For hallucination classification, the word-overlap feature has the most signification impact on the model performance. 

\subsection{Where Does the Model Learn to Hallucinate?}
\label{sec:hallucination_source}
Table \ref{table:dataset} shows that 30\% of the entities in the summaries generated by BART are hallucinated, including 15\% factual hallucinated entities. To generate factual hallucinated entities, the summarization model needs to integrate background knowledge into the summary. One interesting problem is investigate where the model learns that knowledge. Since the BART is pre-trained on a large text corpus and fine-tuned on \textsc{XSum}, the knowledge of hallucinated entities could come from either the pre-training corpus or the \textsc{XSum} training set. To investigate this, we trained a separate CMLM on the CNN/DM dataset.


\begin{figure}[t!]
\centering
\setlength{\belowcaptionskip}{-0.5cm}
\includegraphics[scale=0.50]{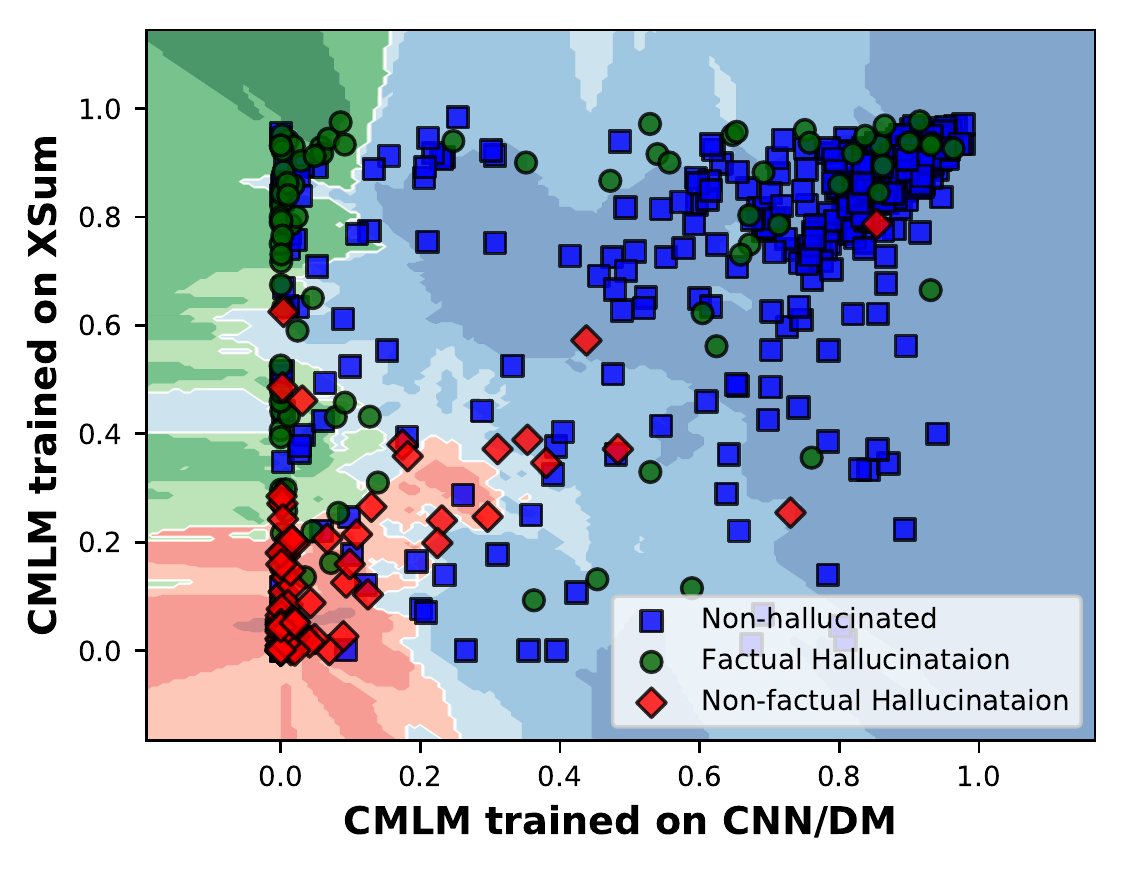}
\caption{Entity distribution over posterior probabilities from $\textsc{CMLM}_{\textsc{XSum}}$ and $\textsc{CMLM}_{\textsc{CNN/DM}}$. The shading shows the classification boundaries of the classifier.}
\label{figure:entity_distribution_2cmlm_nc}
\end{figure}


Figure~\ref{figure:entity_distribution_2cmlm_nc} shows the entity distribution from the two CMLM models. 
For non-hallucinated entities, the distributions are similar; for factual hallucinations, we can find that a large portion of them has very low posterior probabilities under $\textsc{CMLM}_{\textrm{CNN/DM}}$, but high posterior under $\textsc{CMLM}_{\textsc{XSum}}$. This pattern suggests that the knowledge of many factual hallucinations comes from the \textsc{XSum} training set.


We define $\sigma(e_k) = \log \frac{P_{\textrm{CMLM}_\textsc{XSum}}(e_k)}{P_{\textrm{CMLM}_\textsc{CNN/DM}}(e_k)}$. If $\sigma(e_k) \geq 0$, it suggests that $\textsc{CMLM}_{\textsc{XSum}}$ is more confident that $e_k$ is factual than $\textsc{CMLM}_{\textsc{CNN/DM}}$. For a factual hallucination $e_k$, we can infer that the knowledge of $e_k$ is in \textsc{XSum} if $\sigma(e_k)$ is large.
To further verify this, we retrieve the 10 most similar documents from \textsc{\textsc{XSum}} and \textsc{CNN/DM} for each factual hallucinated entity using TF-IDF. Then, we count the number of times each entity appears in those similar training samples. For entities with $\sigma(e_k) \geq 5$, the average number of appearances is 2.19 on \textsc{XSum} and 0.77 on CNN/DM. For entities with $\sigma(e_k) \leq 0$, the average number of appearances becomes 2.85 and 2.46 on \textsc{XSum} and CNN/DM respectively. This further confirms that the knowledge of factual hallucinations with large $\sigma(e_k)$ comes from \textsc{XSum}.

\section{Conclusion}
In this paper, we investigate the hallucination and factuality problems in abstractive summarization. We show that about 30\% of entities generated by state-of-the-art summarization model are hallucinated.
More interestingly, more than half of the hallucinated entities are factual with respect to the source document and world knowledge. 
We propose a novel method based on the entity's prior and posterior probabilities according to masked language models. Our approach outperforms two baseline models on both factuality classification and hallucination detection tasks on human-annotated datasets. In addition, using our classifier as a reward signal vastly improves the factuality of summarization systems.


\bibliographystyle{acl_natbib}
\bibliography{anthology, custom}

\appendix
\clearpage

\section{Appendix}
\subsection{Hallucination Examples}
\label{sec:appeddix_example}
Table \ref{table:example} shows four examples of different classes of hallucinations. In the first example, both entity ``Edinburgh Zoo'' and ``Tian Tian'' are non-hallucinated since they are both mentioned in the source document. In the second example, location ``Cardiff'' is classified as factual hallucination. This location information is not directly inferable from the source document. However, it is factual based on the information we found online. In the third example, the name of the cafe shop ``Waverley'' in the generated summary is hallucinated and non-factual. In the last example, ``Swansea'' is the place where the man is from but not the location of the power station.

\subsection{Experimental Setup}
\label{sec:hyperparameters}
\paragraph{Dataset} We use both \textsc{XSum}, \citet{xsum-emnlp}) and the CNN/Dailymail dataset \citep{NIPS2015_afdec700} in this work. 
CNN/DailyMail is a widely used summarization benchmark with 287,227 training samples, 13,368 validation samples, and 11,490 test samples.
\textsc{XSum} dataset contains 226,711 British Broadcasting Corporation
(BBC) articles. Each article is paired with a single sentence summary written by the BBC journalists. The dataset is split into three subsets: training (204,045, 90\%), validation (11,332, 5\%), and test (11,334, 5\%) sets. 

\paragraph{Language Model Hyperparameters} All language models used in this paper are based on the Transformer encoder-decoder architecture from the Fairseq library \cite{ott2019fairseq} that is written in PyTorch \cite{paszke2017automatic}. For the CMLM training, we initialize the model with the checkpoint of the large BART model. The max sequence length is set to 1024 for both the encoder and decoder modules. We fine-tuned the model for 15,000 steps with the warm-up steps set to 500. We use the standard cross-entropy loss as our objective function with 0.1 label-smoothing \cite{44903}. The Adam optimizer \cite{DBLP:journals/corr/KingmaB14} with $\epsilon=\textrm{1e-8}$ and an initial learning rate 3e-5 are used for training. 
The dropout rate in each layer is set to 0.1. 
All experiments are conducted on 4 Tesla V100 GPUs with 32GB of memory.

\paragraph{RL Training} In the off-line RL experiment, we initialize the model using the BART large model finetuned on \textsc{XSum} dataset\footnote{\url{https://github.com/pytorch/fairseq/tree/master/examples/bart}}. The discount factor $\gamma$ is set to 1 and the learning rate $r$ is set to $1e-5$. We update the model for 30,000 steps in total with 1000 warm-up steps. We use polynomial decay to update the learning rate after each training step. No reward-shaping is used.

To make the training more stable, we use another policy network $\tilde{\pi}_{\theta}$ to compute the importance weight $w$. $\tilde{\pi}_{\theta}$ is kept as a slow copy of $\pi_{\theta}$ with the same model architecture. We use \textit{Polyak updates} to slowly update the weight of $\tilde{\pi}_{\theta}$ in the direction to match $\pi_{\theta}$ every step. The update rate of $\tilde{\pi}_{\theta}$ is set to 0.01.

\subsection{Classification Results on {\datasetname} Dataset}

\label{sec:3class_classification}
\begin{table}[!htbp]
\renewcommand{\arraystretch}{1.1}
\begin{center}
\begin{tabular}{c|ccc}
\toprule
~ & Prec. & Recall & F1 \\ 
\midrule
Non-hallucinated & 97.88 & 92.38 & 95.05 \\
Factual hal. & 60.84 & 84.87 & 70.88 \\
Non-factual hal. & 71.43 & 56.18 & 62.89 \\
\bottomrule
\end{tabular}
\end{center}
\caption{\label{tab:evaluation_three} Evaluation results on {\datasetname}. We report the leave-one-out error of our {\modelname} model with prior, posterior probability and word overlap as features.}
\end{table}

Table \ref{tab:evaluation_three} shows the three-class classification results of our model on {\datasetname} dataset. Since we are the first work (to the best of our knowledge) that distinguishes between factual and non-factual hallucinations, we did not have a baseline model to compare with right now. We compare with other models separately in terms of factuality and hallucination classification in Section \ref{sec:main}.

\begin{table*}[t!]
\small
\renewcommand{\arraystretch}{1.35}
\centering
\begin{tabular}{cm{9cm}m{3.5cm}}
  \toprule
  \bf Category & \bf Source Document & \bf Generated Summary \\
  \hline
  Non-hallucinated & (...) Tian Tian has had cubs in the past in China, before she came on loan to Edinburgh. If she does have a successful delivery, it will be the first time a giant panda has been born in Britain. The panda enclosure at Edinburgh Zoo is due to close to visitors from Saturday ahead of a possible birth. & \hl{Edinburgh Zoo}'s giant panda, \hl{Tian Tian}, could give birth at the end of the month, it has been confirmed. \\ 
  \hline
  \makecell{Factual \\ Hallucination} & The panther chameleon was found on Monday by a dog walker in the wooded area at Marl Park. It had to be put down after X-rays showed all of its legs were broken and it had a deformed spine. RSPCA Cymru said it was an "extremely sad example of an abandoned and neglected exotic pet". (...) & A chameleon that was found in a \hl{Cardiff} park has been put down after being abandoned and neglected by its owners. \\ 
  \hline
  \makecell{Non-factual \\ Hallucination} & The city was brought to a standstill on 15 December last year when a gunman held 18 hostages for 17 hours. Family members of victims Tori Johnson and Katrina Dawson were in attendance. (...) Prime Minister Malcolm Turnbull gave an address saying a "whole nation resolved to answer hatred with love". (...) & Sydney has marked the first anniversary of the siege at the \hl{Waverley} cafe in which two women were killed by a gunman in the Australian city. \\ 
  \hline
  \makecell{Intrinsic \\ Hallucination} & Christopher Huxtable, 34, from Swansea, had been missing since the collapse in February. His body was found on Wednesday and workers who carried out the search formed a guard of honour as it was driven from the site in the early hours of the morning. (...) & The body of a man whose body was found at the site of the \hl{Swansea} Bay Power Station collapse has been removed from the site. \\ 
  \bottomrule
\end{tabular}
\caption{\label{table:example} Examples of four classes of hallucinations.}
\end{table*}

\subsection{Prior/Posterior Probabilities}
\begin{figure*}[ht!]
\setlength{\belowcaptionskip}{-0.5cm}
  \subfloat[\label{fig:entity_classification}]{%
      \includegraphics[width=0.33\textwidth]{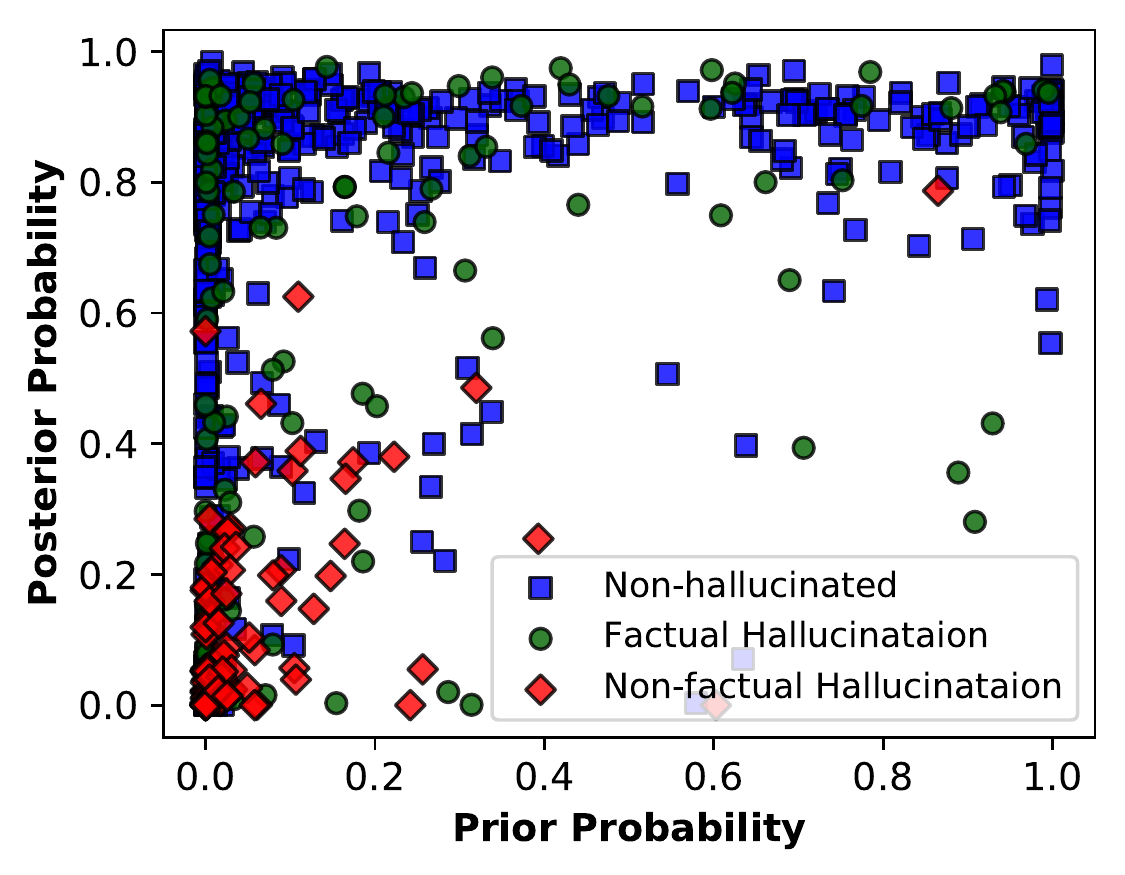}}
\hspace{\fill}
  \subfloat[\label{fig:factuality_classification} ]{%
      \includegraphics[width=0.33\textwidth]{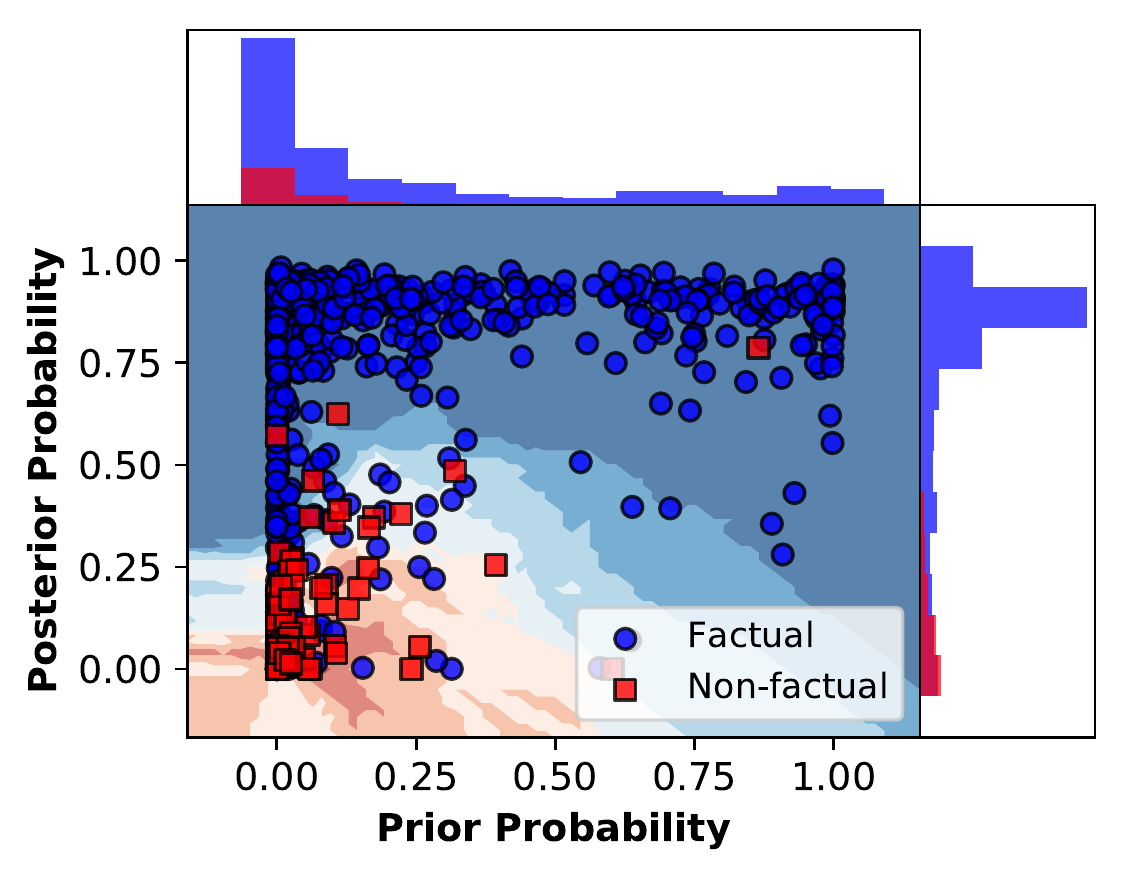}}
\hspace{\fill}
  \subfloat[\label{fig:hallucination_classification}]{%
      \includegraphics[width=0.33\textwidth]{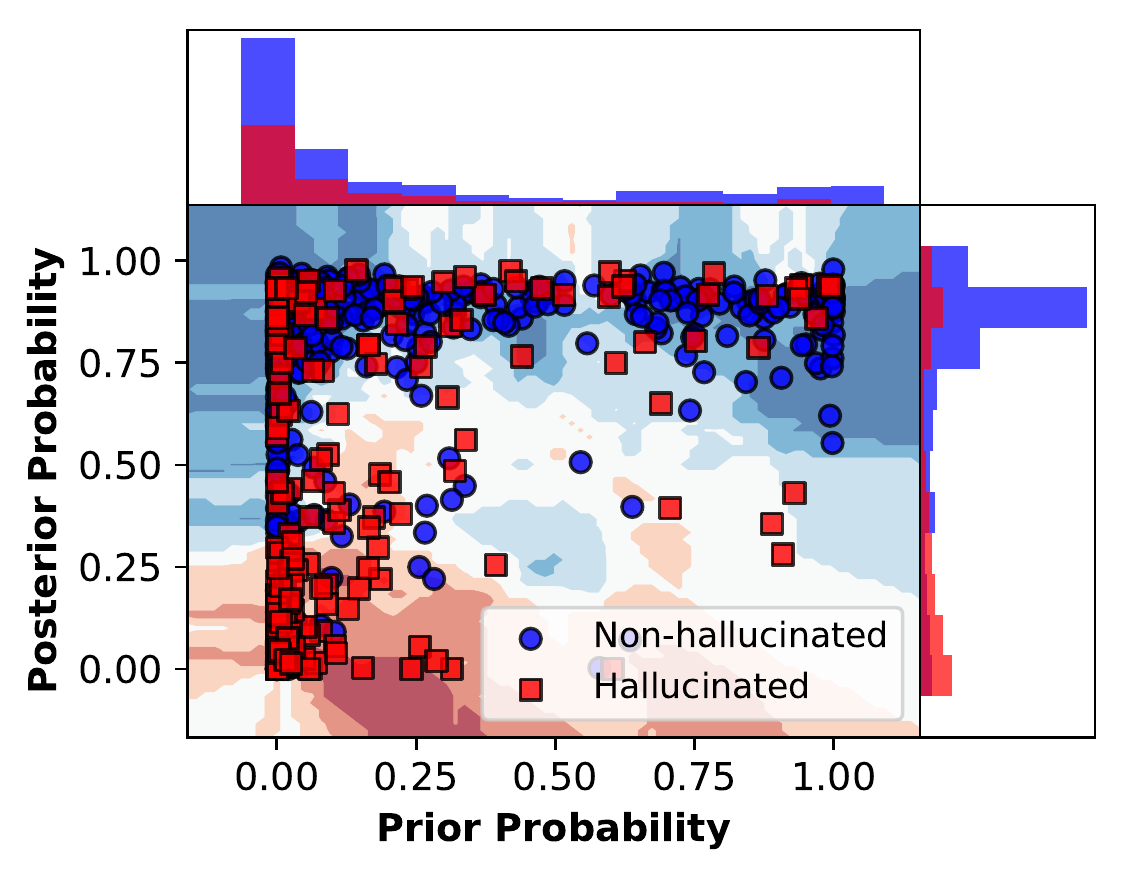}}\\
\caption{\label{fig:distribution} The distribution of entities over prior/posterior probability. Each point in the figure represents an entity $(p_{\textrm{prior}}(e_k), p_{\textrm{pos}}(e_k))$ and shading indicates the confidence of the classifier. (a) The distribution of entities; (b) The entity factuality classification results with KNN ($k=20$) classifier. Both factual hallucinated and non-hallucinated entities are colored blue; (c) The KNN ($k=20$) classification boundaries of hallucinated and non-hallucinated entities.}
\end{figure*}

Figure~\ref{fig:distribution} plots entities in the {\datasetname} dataset according to their prior and posterior probabilities and shows the KNN classification boundaries of {\modelname} w/o overlap.
In Figure~\ref{fig:entity_classification}, we find that the non-factual hallucinated entities are clustered around the origin. This is in line with our expectations since non-factual hallucinations have lower prior and posterior probabilities. Both factual hallucinated and non-hallucinated entities are gathered in the top area with high posterior probabilities.

In Figure~\ref{fig:factuality_classification}, the KNN classifier separates the factual and non-factual entities with clear boundaries. A large part of the factual hallucinated entities are correctly identified by $\textrm{CMLM}_\textsc{XSum}$ with relatively high posterior probabilities. This explains our model's superior performance on factuality checking.
The top and right histograms in Figure \ref{fig:factuality_classification} show the entity distribution over prior and posterior probability value respectively.
As shown in \ref{fig:factuality_classification}'s histogram, factual entities have significantly higher posterior probability than that of non-factual entities on average.

\begin{figure}[t!]
\centering
\setlength{\belowcaptionskip}{-0.3cm}
\includegraphics[scale=0.40]{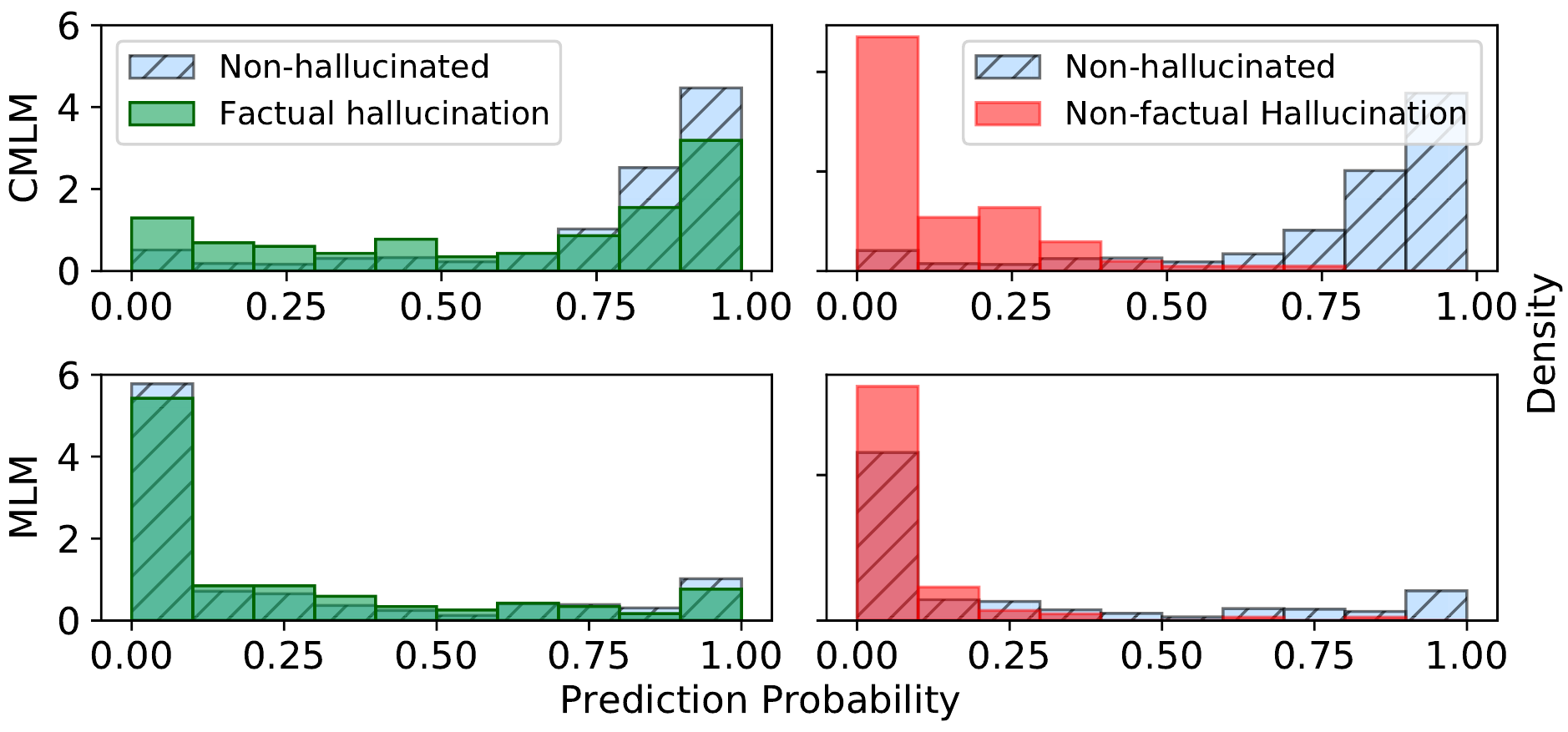}
\caption{Normalized histogram of model prediction probability for three classes of entities. The first row shows the entities' posterior probability calculated using CMLM. The second row shows the prior probability from MLM.}
\label{figure:hist}
\end{figure}
Figure~\ref{figure:hist} shows histograms of the prior and posterior probabilities of entities from MLM and $\textrm{CMLM}_{\textsc{XSum}}$, separated by their class (i.e., whether they are hallucinated and/or factual). Non-hallucinated entities have higher posterior probability than factual and non-factual hallucinations on average. The average posterior probability for non-hallucination, factual hallucinations, and non-factual hallucinations are 0.763, 0.599, and 0.133 respectively.

\subsection{Evaluating Entity Factuality on Noisy Training Data}
\label{sec:noisy_analysis}

\begin{figure}[t!]
\centering
\setlength{\belowcaptionskip}{-0.5cm}
\includegraphics[scale=0.37]{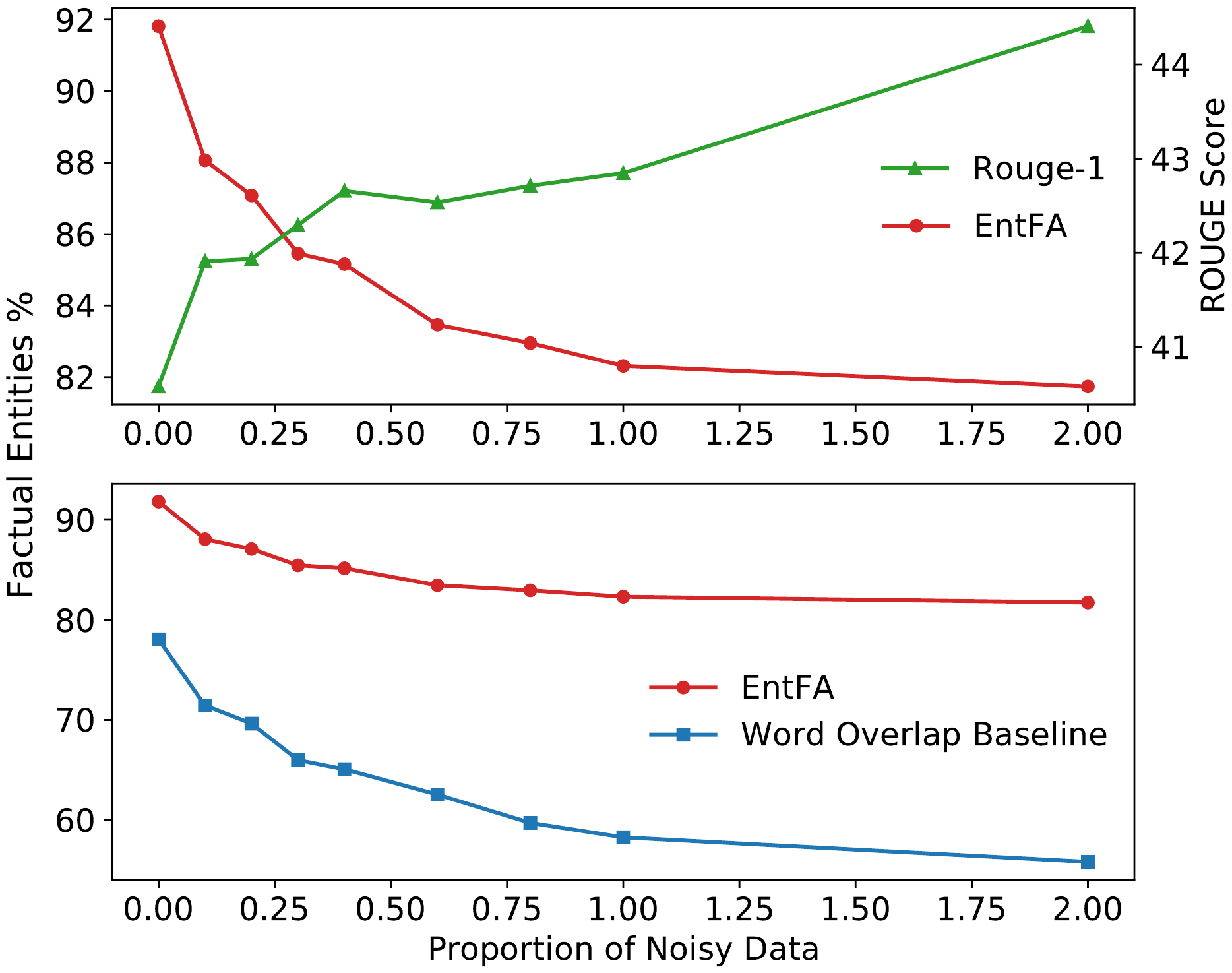}
\caption{Evaluation of an abstractive summarization model (BART) trained on datasets with different levels of noise. The y-axis on the left represents the percentage of factual entities classified as factual by ({\modelname}) or the word overlap baseline. The y-axis on the right indicates ROUGE-1 scores. X-axis $=0$ and x-axis $=1.0$ means that the model is trained on 50k clean samples and 50k noisy samples respectively; x-axis $=0.5$ represents the model trained on a mix of 25k clean samples and 25k noisy samples. X-axis $=2.0$ represents a model that is trained on 100k noisy samples. All models are tested on \textsc{XSum}'s official test set. We observe a similar trend with the \textsc{PEGASUS} model (Figure \ref{figure:noise_pegasus}).}
\label{figure:noise}
\end{figure}

Recent work \cite{narayan2021planning, nan-etal-2021-entity} has shown that filtering out noisy training samples in the \textsc{XSum} dataset can mitigate the hallucination issue. Therefore, we divide the XSum training set into clean samples and potentially noisy samples. Potentially noisy samples are samples where the reference summary contains entities that does not appear in the source. This gives us around 150k potentially noisy training samples and 50k clean training samples. Then, we mix the clean samples with noisy samples at different proportions to create training sets with different levels of noise. Figure~\ref{figure:noise} shows the evaluation results of summarization models trained on these datasets. We can see that the model generates fewer factual entities as the training set gets noisier. Also, it shows that ROUGE score is not a favorable metric in terms of factuality evaluation. Since with the training set size fixed, the model seems to achieve higher ROUGE score at the expense of entity factuality. In addition, this indicates that if the system is optimized only for ROUGE, they may inadvertently harm factual consistency.

We also observe that the word overlap method predicts much lower entity factuality rate than {\modelname}. This is due to the fact that the word overlap method cannot identify factual hallucinations and introduce many false-negative samples. To verify this,
we extracted all entities from summaries generated by the model trained on 50k noisy samples (x-axis $=1.0$).
Among these entities, there are 7,358 entities that do not appear in the source but are predicted as factual by our model.  We find that 50.5\% of these entities can be found in the reference summary. As a contrast, only 12.7\% entities predicted as non-factual by our model can be found in the reference. 

\label{sec:pegasus}
\begin{figure}[!htbp]
\centering
\includegraphics[scale=0.4]{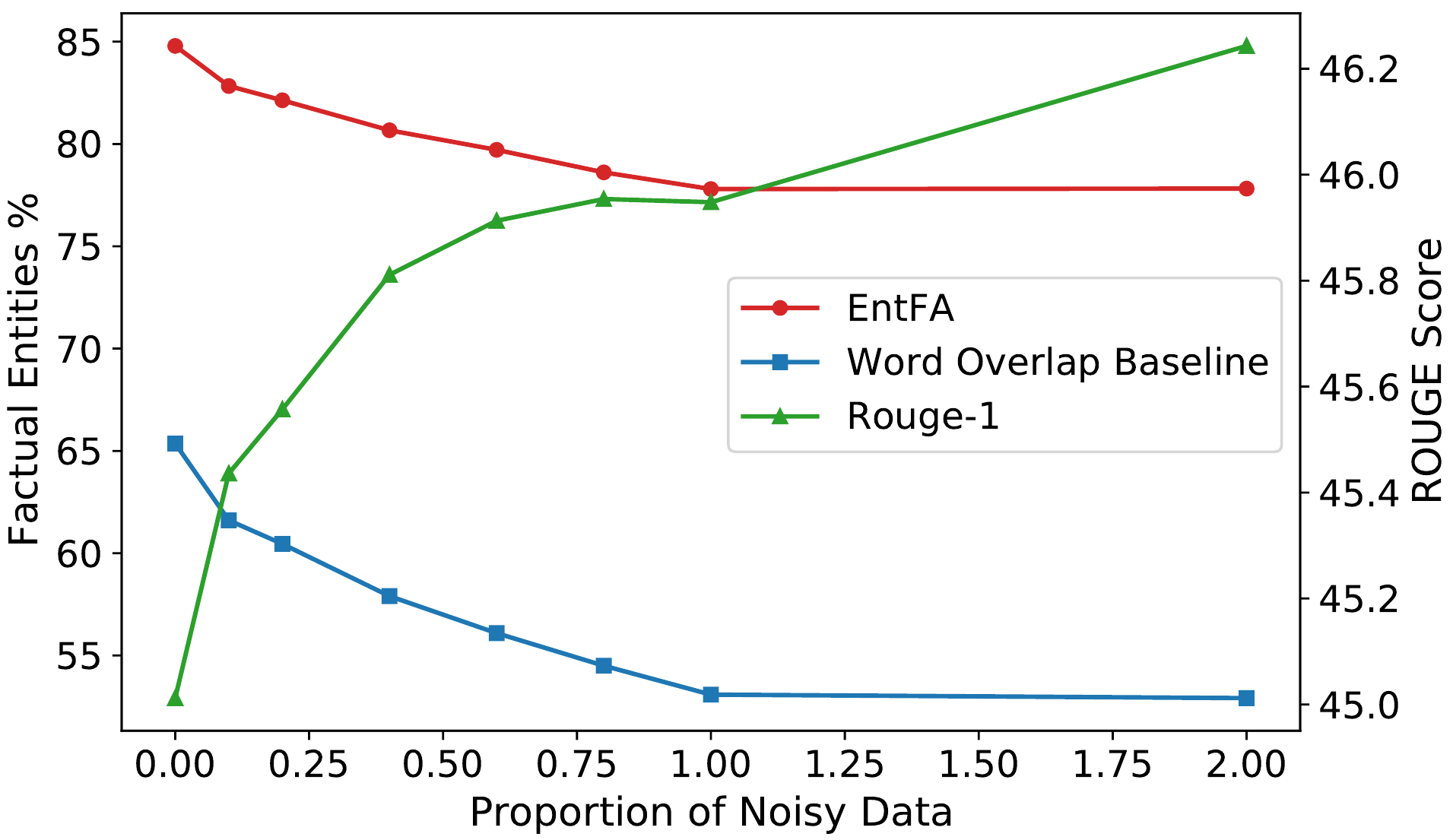}
\caption{Evaluation of $\textsc{PEGASUS}_{\textsc{LARGE}}$ trained on datasets with different levels of noises.}
\label{figure:noise_pegasus}
\end{figure}

Figure \ref{figure:noise_pegasus} shows the evaluation result of \textsc{PEGASUS} model \cite{pmlr-v119-zhang20ae} follows the evaluation set up in Section \ref{sec:noisy_analysis}. Both figures show a similar trend that the models get higher ROUGE score when trained on noisier dataset with the cost of generating more non-factual entities.

Compared with BART model, \textsc{PEGASUS} generates more hallucinated entities and has higher ROUGE score overall. For instance, when both trained on 50k clean data, \textsc{PEGASUS} has ROUGE-1 score 0.450 compared with BART's 0.406. The predicted factual entity rate for \textsc{PEGASUS} and BART is 84.79\% and 91.81\% respectively. This may be due to the fact that \textsc{PEGASUS} is pre-trained on a much larger corpus than BART. We leave the study of this phenomenon to future work.

\subsection{Why not Use CLM?}
\label{sec:compare_clm_cmlm}
\citet{filippova-2020-controlled}'s work on data-to-text generation shows that low posterior probability from a CLM during decoding indicates hallucination.
Take the summarization model as an example, if an entity is generated with very low posterior probability, it is likely that the generated entity is hallucinated and non-factual. However, compared with CMLM, CLM has more uncertainty during decoding since the right context of the entity is not determined. The uncertainty of the CLM comes from both content selection (text content and
structure) and lexical choice \cite{xu-etal-2020-understanding-neural}. For CMLM though, the uncertainty is mostly reduced to the latter. 

\begin{figure}[!htbp]
\centering
\setlength{\belowcaptionskip}{-0.4cm}
\includegraphics[scale=0.55]{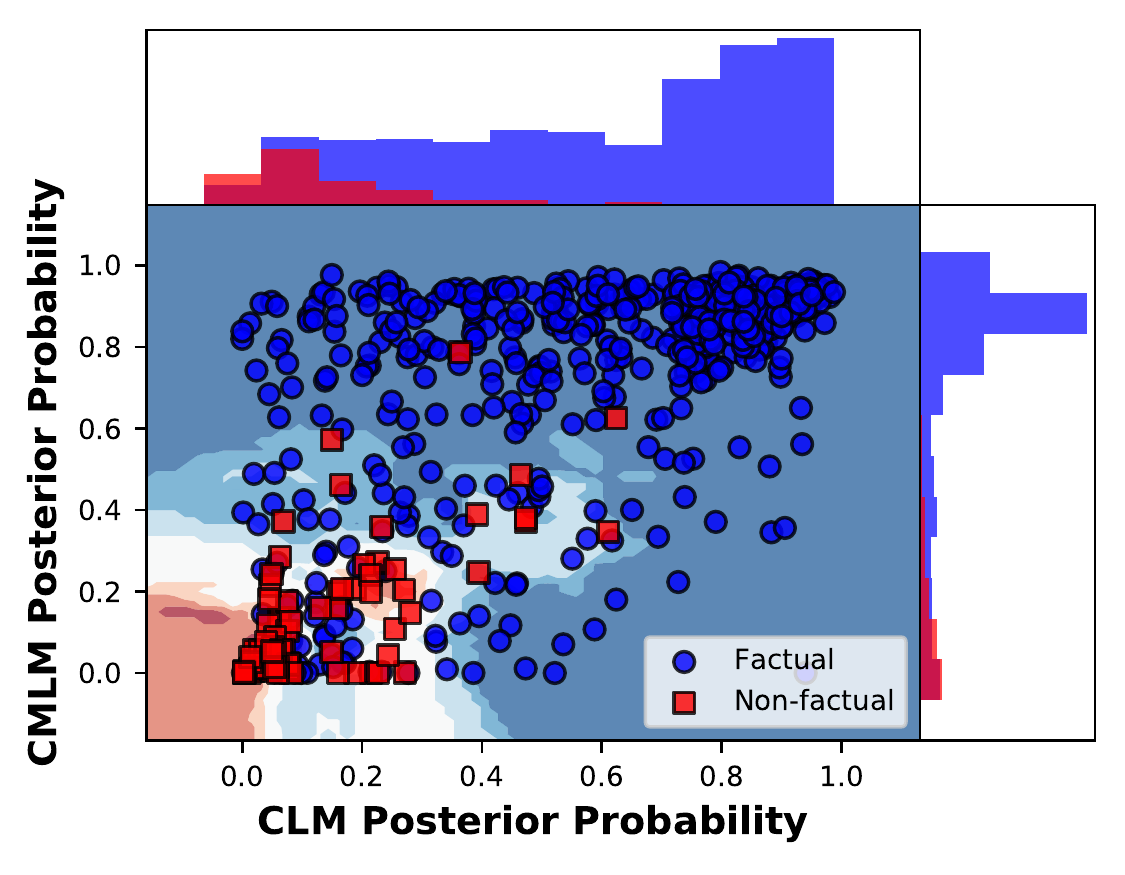}
\caption{Posterior probabilities calculated from CLM and CMLM. Both models are trained on \textsc{XSum} dataset.}
\label{figure:clm_cmlm}
\end{figure}

\begin{figure}[!htbp]
\centering
\setlength{\belowcaptionskip}{-0.4cm}
\includegraphics[scale=0.52]{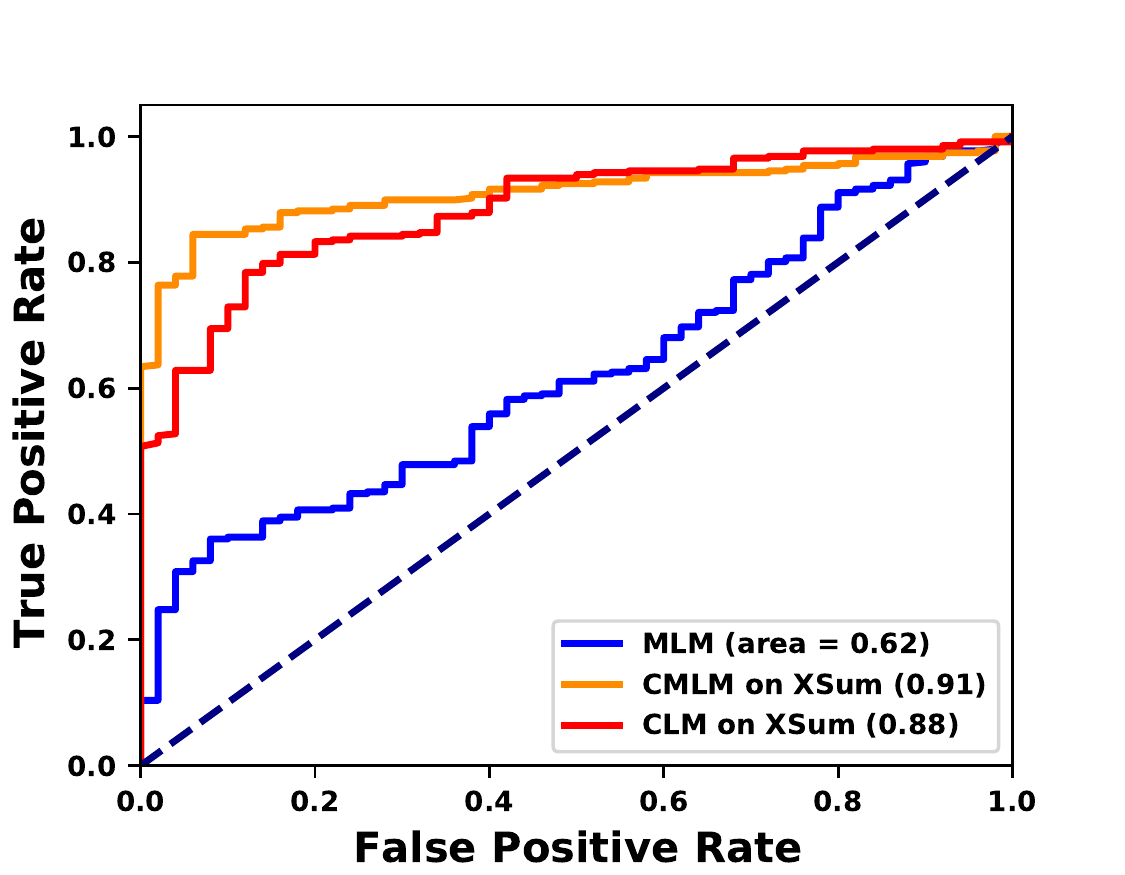}
\caption{ROC curve of entity's posterior probability and factuality.}
\label{figure:auc_clm_cmlm}
\end{figure}

Figure \ref{figure:clm_cmlm} show the entity posterior probabilities from CLM and CMLM model. As shown in the figure, we can find that most factual entities (blue points) are above the $x=y$ line. This means CMLM gives more certainty to the same factual entity than CLM. The ROC curve in Figure \ref{figure:auc_clm_cmlm} further shows this. As the lines get closer to the origin, the threshold becomes larger, and CMLM has a higher TPR than CLM. This means CMLM will classify more entities as factual. The higher AUC value of CMLM further demonstrates that CMLM is a better choice for factuality checking than CLM.

\end{document}